\documentclass[letterpaper, 10 pt, conference]{ieeeconf}  

\usepackage{graphicx}
\usepackage{amsmath}
\usepackage{amssymb}
\usepackage{stfloats}

\usepackage{xcolor}
\usepackage{url}
\usepackage{tensor}
\usepackage{mathtools}
\usepackage{amsmath}
\usepackage{bbm}
\usepackage{enumerate}
\usepackage[ruled,linesnumbered]{algorithm2e}
\usepackage{siunitx}
\usepackage{tabularx}
\usepackage{multirow}
\usepackage{hhline}
\newcommand{\tabincell}[2]{\begin{tabular}{@{}#1@{}}#2\end{tabular}}

\newcommand{\red}{\textcolor{red}}

\IEEEoverridecommandlockouts                              

\title{\LARGE \bf
Automatic Recognition of Abdominal Organs in Ultrasound Images based on Deep Neural Networks and K-Nearest-Neighbor Classification 
}

\author{Keyu Li, Yangxin Xu, and Max Q.-H. Meng$^{*}$, \textit{Fellow}, \textit{IEEE}
\thanks{This work was partially supported by National Key R\&D program of China with Grant No. 2019YFB1312400, Hong Kong RGC GRF grant \#14210117, Hong Kong RGC TRS grant T42-409/18-R  and Hong Kong RGC GRF grant \#14211420 awarded to Max Q.-H. Meng.}
\thanks{K. Li and Y. Xu are with the Department of Electronic Engineering, The Chinese University of Hong Kong, Hong Kong, China (e-mail: kyli@link.cuhk.edu.hk; yxxu@link.cuhk.edu.hk).}%
\thanks{Max Q.-H. Meng is with the Department of Electronic and Electrical Engineering of the Southern University of Science and Technology in Shenzhen, China, on leave from the Department of Electronic Engineering, The Chinese University of Hong Kong, Hong Kong, and also with the Shenzhen Research Institute of the Chinese University of Hong Kong in Shenzhen, China (e-mail: max.meng@ieee.org).}%
\thanks{$^{*}$Corresponding author.}%
}

\begin{document}

\maketitle
\thispagestyle{empty}
\pagestyle{empty}

\begin{abstract}
Abdominal ultrasound imaging has been widely used to assist in the diagnosis and treatment of various abdominal organs. In order to shorten the examination time and reduce the cognitive burden on the sonographers, we present a classification method that combines the deep learning techniques and k-Nearest-Neighbor (k-NN) classification to automatically recognize various abdominal organs in the ultrasound images in real time. Fine-tuned deep neural networks are used in combination with PCA dimension reduction to extract high-level features from raw ultrasound images, and a k-NN classifier is employed to predict the abdominal organ in the image. We demonstrate the effectiveness of our method in the task of ultrasound image classification to automatically recognize six abdominal organs. A comprehensive comparison of different configurations is conducted to study the influence of different feature extractors and classifiers on the classification accuracy. Both quantitative and qualitative results show that with minimal training effort, our method can ``lazily" recognize the abdominal organs in the ultrasound images in real time with an accuracy of $96.67\%$. Our implementation code is publicly available at \url{https://github.com/LeeKeyu/abdominal_ultrasound_classification}.
\end{abstract}
\section{INTRODUCTION}

Medical ultrasound imaging uses high-frequency acoustic waves to generate B-mode images of internal body structures, such as muscles, nerves and vessels. Due to the advantages of portability, non-invasiveness, low cost and real-time capabilities over other medical imaging techniques, ultrasound imaging has been widely accepted as both a diagnostic and a therapeutic tool in various medical disciplines, such as cardiology, urology, neurology, obstetrics and gynecology \cite{ultrasound}. In routine ultrasound examinations, a trained sonographer manually positions an ultrasound transducer on the patient skin surface and navigate it towards the correct imaging plane that can provide a clear and diagnostically valuable view of the target organ. 
Abdominal ultrasound imaging is a safe and painless imaging technique that is widely used to assist the diagnosis of abdominal pain or distention, abnormal liver function, enlarged abdominal organ, kidney stones, gallstones, and the abdominal aortic aneurysm (AAA) \cite{abdominal}. Besides, abdominal ultrasound imaging is also frequently performed to provide guidance for biopsies in the abdominal region. Various internal organs can be imaged during the abdominal ultrasound imaging, such as the kidneys, liver, gallbladder, bile ducts, pancreas, spleen and abdominal aorta.

\begin{figure}[t]
      \centering
      \includegraphics[scale=1.0,angle=0,width=0.49\textwidth]{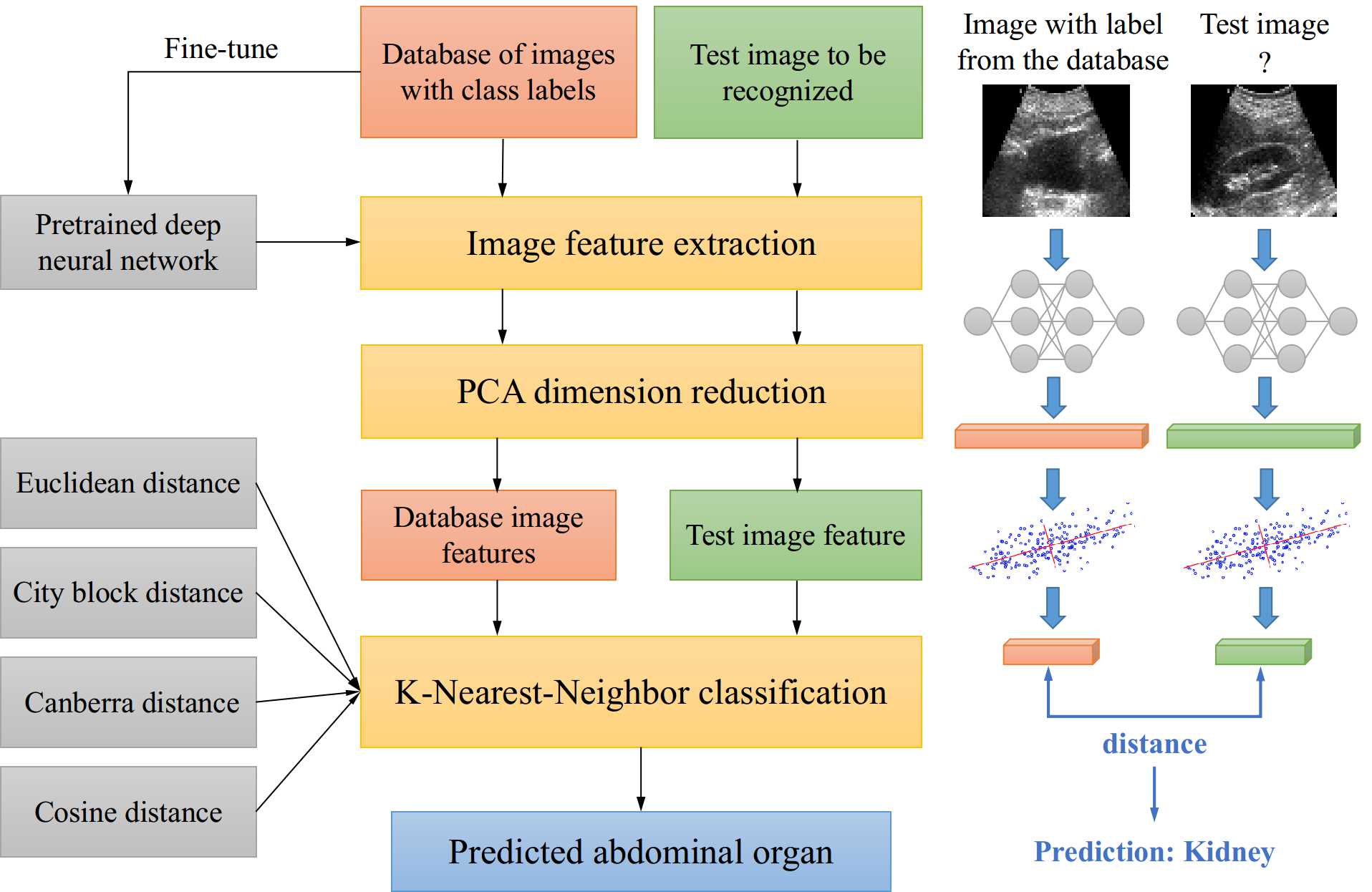}
      \caption{The overall workflow of the proposed method for abdominal organ recognition from ultrasound images.}
      \label{workflow}
\end{figure}

However, in routine ultrasound examinations, the manual positioning of the probe is usually tedious and time-consuming, as the clinician has to continuously tilt, slide, and rotate the ultrasound transducer based on the acquired ultrasound images to search for an anatomy of interest \cite{9399640}\cite{li2021autonomous}. Also, the imaging quality is highly dependent on the knowledge and skills of the sonographer. Moreover, the interpretation of the highly dynamic ultrasound images will bring information overload and high cognitive burdens on the sonographer  \cite{berg2006operator}. Especially during abdominal ultrasound examinations, since various internal organs are captured by the ultrasound probe during the scan, the clinician has to manually identify multiple abdominal organs from the ultrasound images to correctly position the transducer. 

To this end, an automatic recognition of different abdominal organs would be helpful to make the ultrasound imaging process easier and faster. Some researchers proposed to automatically localize the epigastric region based on some surface landmarks (e.g., the umbilicus and mammary papillae) captured by an external camera \cite{6610635}. However, this method can only roughly locate the abdominal region, and the internal organs are difficult to be accurately identified with the surface landmarks. Since the ultrasound image content provides rich information of the internal anatomical structures, we envision a robotic system that can automatically recognize and annotate the internal abdominal organs in the ultrasound image during the acquisition, so that the doctor will only need to make a simple decision to move the transducer based on the spatial relationship between different organs to quickly and accurately reach the target organ. This technology can also be integrated in a robotic ultrasound system to better assist the tele-operated or autonomous ultrasonography, where a robotic manipulator is controlled by an off-site clinician to acquire images on the patients \cite{9399640}\cite{optimization}\cite{confimap}. Therefore, the automatic classification of abdominal ultrasound images holds great promise to reduce the workload of the sonographers, shorten the examination time, and reduce the user-dependency of the ultrasound imaging results.

\begin{figure}[t]
      \centering
      \includegraphics[scale=1.0,angle=0,width=0.42\textwidth]{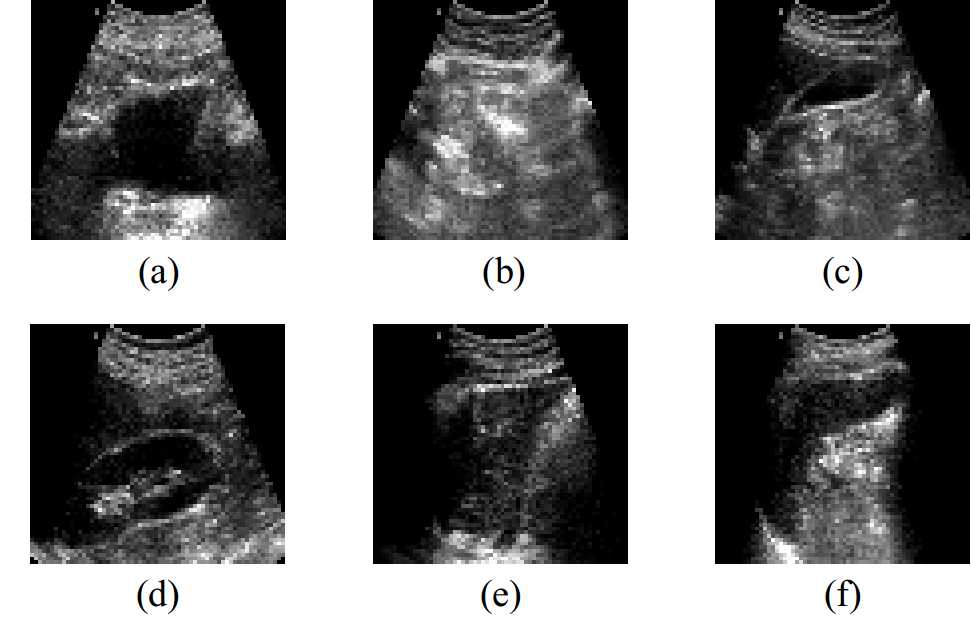}
      \caption{Abdominal ultrasound images of 6 different organs (a) bladder, (b) bowel, (c) gallbladder, (d) kidney, (e) liver, and (f) spleen.}
      \label{organs}
\vspace{-0.1cm}
\end{figure}

With the rapid development in computer vision and machine intelligence, a large number of deep learning-based methods have been proposed to analyze the medical ultrasound images in a data-driven manner, yielding frontier results in the classification, detection, and segmentation tasks for different anatomical structures \cite{aiusanalysis}. Most of existing studies on ultrasound image classification have been focused on the diagnosis of a specific organ, such as the detection of tumors, lesions or nodules \cite{guo2017ceus}\cite{ma2017pre}. Only a few methods have been proposed for classification of several abdominal organs in the ultrasound images \cite{cheng2017transfer}\cite{xu2018less}. However, since these methods directly apply deep neural networks (NNs) for classification, they use a large amount of well-annotated ultrasound data to train the models, which is difficult and expensive to collect \cite{ma2017pre}. 

Therefore, we propose a method that combines deep learning with the k-Nearest-Neighbor (k-NN) classification to recognize multiple abdominal organs in the ultrasound images in real time. The overall workflow of our method is illustrated in Fig. \ref{workflow}. Specifically, our method employs fine-tuned deep NNs and PCA dimension reduction for feature extraction from ultrasound images, and uses a k-NN classifier to recognize the abdominal organs in the image.
Deep NNs have superior capability of extracting high-level features from raw image data but require large-scale training data to achieve a good performance. Instead, the k-NN classifier is a lazy and non-parametric algorithm that has the good characteristics of being simple and easy to use, and has a reasonable accuracy \cite{altman1992introduction}. Combining the advantages of the two methods, we demonstrate that we can achieve good classification performance with minimal training effort using a relatively small ultrasound dataset.

\section{METHOD}

\subsection{Data Pre-processing}
A total of 360 ultrasound images of 6 abdominal organs, namely, the bladder, bowel, gallbladder, kidney, liver, and spleen, are obtained from a publicly available dataset of abdominal ultrasound images \cite{dataset}. Exemplary images are shown in Fig. \ref{organs}(a-f). The dataset is randomly split into a training set (or database) of 300 images (50 images per organ) and a test set of 60 images (10 images per organ) in an image-wise manner. Each image is resized to $64 \times 64$ pixels and converted into 24-bit RGB format to serve as input to the feature extractors.

\subsection{Fine-tuning of Deep Neural Networks for Feature Extraction}
In recent years, a large number of deep NN architectures have been proposed for image classification and successfully applied on medical images \cite{litjens2017survey}. Among the state-of-the-art models, we select the ResNet \cite{he2016deep} and DenseNet \cite{huang2017densely} architectures with similar model size to achieve a balance between complexity and accuracy. Three versions of each model with different depths are used for a comparison, namely, the ResNet-50, ResNet-101, ResNet-152, and DenseNet-121, DenseNet-169 and DenseNet-201. We implement the deep NN based feature extraction in Python based on the Keras API for TensorFlow 2.0 \cite{keras}. 

\subsubsection{Pretrained Networks for Feature Extraction}
We first forgo the top fully connected (FC) layer of each NN architecture, and use publicly available weights for the six deep NN models pretrained on the ImageNet dataset \cite{deng2009imagenet}. Since the top layer for classification is removed, the pretrained NNs are used to extract high-level features of the input ultrasound images, and output feature vectors of dimension 2048, 2048, 2048, 1024, 1664 and 1920 for each deep NN architecture. The output features are then normalized by

\begin{equation}
x' = \frac{x-x_{min}}{x_{max}-x_{min}}
\label{normalize}
\end{equation}

\begin{figure}[t]
      \centering
      \includegraphics[scale=1.0,angle=0,width=0.49\textwidth]{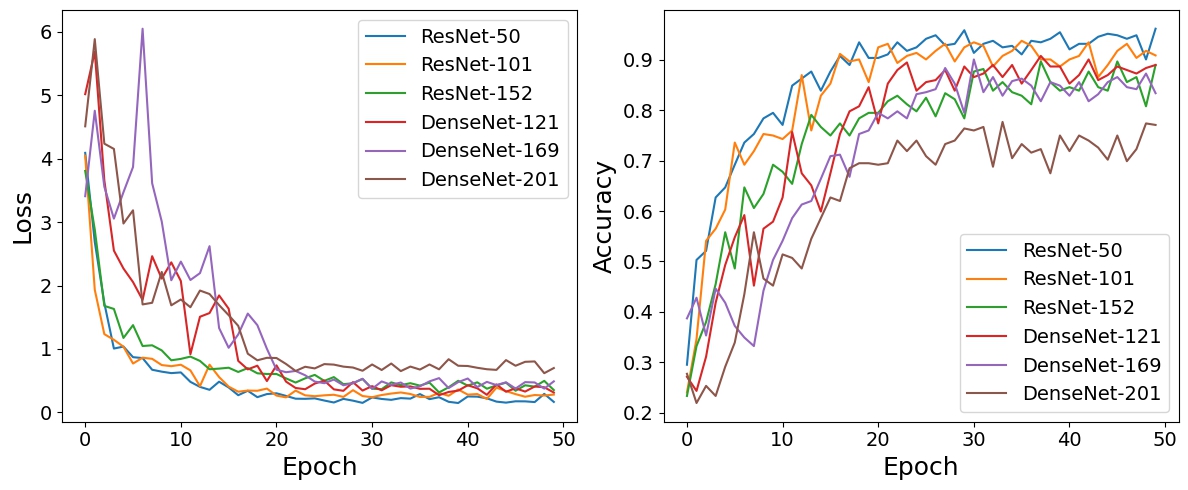}
      \caption{Learning curves of the ResNet-50, ResNet-101, ResNet-152, DenseNet-121, DenseNet-169 and DenseNet-201 models during fine-tuning on our data for abdominal organ classification, which indicate the (a) training loss and (b) training accuracy as functions of the training epochs.}
      \label{curve}
\end{figure}

\subsubsection{Fine-tuned Networks for Feature Extraction}
Then, we add an FC layer with 6 outputs corresponding to 6 abdominal organs to each pretrained NN model, and initialize them with random weights. Each NN model is fine-tuned on our training data using stochastic gradient descent and cross-entropy loss with a batch size of 8 for 50 epochs to achieve stable performance. The learning rate is initialized as 0.01 and reduced by 0.2 when the validation loss stagnates for 3 epochs. The learning curves of each NN architecture are shown in Fig. \ref{curve}. After fine-tuning, we forgo the final FC layers are removed and the deep NN models are used as the feature extractor, and the resulting features are also normalized using (\ref{normalize}).

\subsubsection{Dimension Reduction with PCA}
In order to extract important information from the original features generated by the deep NNs, we incorporate the principal component analysis (PCA) technique for dimension reduction of the output features of each deep NN model. The features of the images in the database are first normalized to have zero-mean to calculate the principal components. We select the number of components such that they account for 99\% variance in the database. During inference, the same PCA transform is applied to the test images after subtracting the mean of the database images. 

\subsection{Distance Metrics for Similarity Measurement}
In order to compare the similarity between the test image and images in the database, we calculate the distances between the extracted feature vectors using the following four distance metrics.

\subsubsection{Euclidean Distance}
Euclidean distance is also known as L2 norm or Ruler distance, which is calculated by the root of the sum of the square of differences between the opposite values in two vectors:
\begin{equation}
    d(x,y) = \sqrt{\sum_{i=1}^{n} (x_i-y_i)^2}
\end{equation}
where n is the dimension of the feature vectors $x$ and $y$.

\subsubsection{City Block Distance}
City block distance, also known as Manhattan distance or L1 norm, is calculated by the sum of the absolute differences between the opposite values in two vectors, which is less sensitive to large differences than Euclidean distance:

\begin{equation}
    d(x,y) = \sum_{i=1}^{n} |x_i-y_i|
\end{equation}
where n is the dimension of the feature vectors $x$ and $y$.

\subsubsection{Canberra Distance}
Canberra distance, which is introduced by \cite{lance1966computer}, is a weighted version of City block distance, where the absolute difference between the attribute values of the two vectors is divided by the sum of the absolute attribute values before the summing:

\begin{equation}
    d(x,y) = \sum_{i=1}^{n} \frac{|x_i-y_i|}{|x_i| + |y_i|}
\end{equation}
where n is the dimension of the feature vectors $x$ and $y$. As indicated by the definition of Canberra distance, this metric is sensitive to small changes near zero, which means proportional differences will have a larger influence on the value of this distance metric than absolute differences. 

\subsubsection{Cosine Distance}
The Cosine distance is derived from the Cosine similarity, which measures the angle between two vectors:

\begin{equation}
    d(x,y) = 1- \frac{\sum_{i=1}^{n} x_i y_i}{\sqrt{\sum_{i=1}^{n} x_i^2 \sum_{i=1}^{n} y_i^2}}
\end{equation}
where n is the dimension of the feature vectors $x$ and $y$. Since the Cosine distance measures the differences in directions of two vectors, it is not influenced by the magnitude of the vectors.

\subsection{K-Nearest-Neighbor Classification}

The k-Nearest-Neighbor (k-NN) approach classifies a test sample with an unknown label based on the majority of the most similar samples among its k nearest neighbors in the database \cite{altman1992introduction}. Given a test feature vector, the algorithm first computes the distance to all the training data in the database using a specific distance metric, then k nearest neighbors are identified, and the class labels of the k nearest neighbors are used to determine the class label of the unknown input data by taking the majority vote. 

In our task, the database contains extracted feature vectors of ultrasound images with six classes (different abdominal organs). In order to classify which organ is in the test image, we first identify the k nearest neighbors to the test image feature in the database according to the distance metrics discussed in Section II-C, and then predict the the organ class by taking the majority vote result among the classes of the k neighbors. The value of k is chosen as $k \in \{1,3,5,7,9\}$.

\section{EXPERIMENTS}

\begin{table*}[tb] \renewcommand\arraystretch{1.2} \small
\centering
\caption{Accuracy of Abdominal Organ Recognition in Ultrasound Images using Different Feature Extractors and Classifiers}
\begin{tabular}{|m{0.1\textwidth}<{\centering}|m{0.15\textwidth}<{\centering}||m{0.1\textwidth}<{\centering}|m{0.1\textwidth}<{\centering}|m{0.1\textwidth}<{\centering}|m{0.1\textwidth}<{\centering}||m{0.1\textwidth}<{\centering}|m{0.1\textwidth}|}
\hline
\multicolumn{2}{|c||}{\multirow{4}{*}{\tabincell{c}{Deep NN Feature Extractor}}}&\multicolumn{5}{c|}{Classification Accuracy} \\
\cline{3-7}
\multicolumn{2}{|c||}{} & \multicolumn{4}{c||}{K-NN Classifier} & \multirow{3}{*}{\tabincell{c}{FC Layer  \\ Classifier}}\\
\cline{3-6}
\multicolumn{2}{|c||}{} & Euclidean distance& City block distance & Canberra distance& Cosine distance&  \\
\hhline{|=|=|=|=|=|=|=|}
\multirow{3}{*}{ResNet-50} & Pretrained & $71.67\%$ & $76.67\%$ & $83.33\%$ & $80.00\%$ & --\\
\cline{2-7}
& Fine-tuned w/o PCA  & $93.33\%$ & $\mathbf{95.00}\%$ & \red{$\mathbf{96.67\%}$} & \red{$\mathbf{96.67\%}$} & \red{$\mathbf{96.67\%}$}\\
\cline{2-7}
& Fine-tuned w/ PCA & $\mathbf{93.33}\%$ & $93.33\%$ & $95.00\%$&$95.00\%$ & --\\
\hhline{|=|=|=|=|=|=|=|}
\multirow{3}{*}{ResNet-101} & Pretrained & $86.67\%$ &  $86.67\%$ &  $81.67\%$ &  $85.00\%$ & --
\\
\cline{2-7}
& Fine-tuned w/o PCA  & $90.00\%$ & $90.00\%$ & $90.00\%$&${91.67\%}$& $90.00\%$\\
\cline{2-7}
& Fine-tuned w/ PCA & $\mathbf{90.00}\%$ & $\mathbf{90.00}\%$ & $\mathbf{91.67}\%$&$\mathbf{91.67\%}$&  --\\
\hhline{|=|=|=|=|=|=|=|}
\multirow{3}{*}{ResNet-152} & Pretrained &  $83.33\%$ & $83.33\%$ &   $83.33\%$ & $81.67\%$& --
\\
\cline{2-7}
& Fine-tuned w/o PCA  & $91.67\%$ & $91.67\%$ & $90.00\%$ &$90.00\%$ & $80.00\%$ \\
\cline{2-7}
& Fine-tuned w/ PCA & $\mathbf{91.67}\%$ & $\mathbf{95.00\%}$ & $\mathbf{91.67}\%$&$\mathbf{90.00}\%$&  --\\
\hhline{|=|=|=|=|=|=|=|}
\multirow{3}{*}{DenseNet-121} & Pretrained &  $81.67\%$ & $85.00\%$ & $85.00\%$ & $80.00\%$ & --
\\
\cline{2-7}
& Fine-tuned w/o PCA  &$95.00\%$ & $93.33\%$ & $95.00\%$ &$93.33\%$& $91.67\%$\\
\cline{2-7}
& Fine-tuned w/ PCA & $\mathbf{95.00}\%$ & \red{$\mathbf{96.67\%}$} & $\mathbf{95.00}\%$&$\mathbf{95.00}\%$& --\\
\hhline{|=|=|=|=|=|=|=|}
\multirow{3}{*}{DenseNet-169} & Pretrained & $\mathbf{91.67\%}$ & $86.67\%$ & $88.33\%$ & $88.33\%$ & --
\\
\cline{2-7}
& Fine-tuned w/o PCA  & $90.00\%$ & $90.00\%$ & $90.00\%$&$90.00\%$& $88.33\%$ \\
\cline{2-7}
& Fine-tuned w/ PCA & $90.00\%$ & $\mathbf{90.00}\%$ &$\mathbf{90.00}\%$ &$\mathbf{90.00}\%$ & -- \\
\hhline{|=|=|=|=|=|=|=|}
\multirow{3}{*}{DenseNet-201} & Pretrained & $85.00\%$ & $83.33\%$ & $83.33\%$ & $83.33\%$ & --
\\
\cline{2-7}
& Fine-tuned w/o PCA & $85.00\%$ & $81.67\%$ & $85.00\%$&$\mathbf{88.33\%}$& $73.33\%$\\
\cline{2-7}
& Fine-tuned w/ PCA & $\mathbf{86.67}\%$ & $\mathbf{83.33\%}$ & $\mathbf{85.00}\%$&$83.33\%$&  --\\
\hline
\end{tabular}
\label{result}
\end{table*}

In order to validate the effectiveness of our method, we conduct both quantitative and qualitative experiments to evaluate the performance of our method in the recognition of six abdominal organs from the ultrasound images.

\subsection{Quantitative Evaluation}
We first quantitatively evaluate the classification accuracy of our method when using different feature extractors, distance metrics and classifiers. 

\begin{figure}[t]
      \centering
      \includegraphics[scale=1.0,angle=0,width=0.49\textwidth]{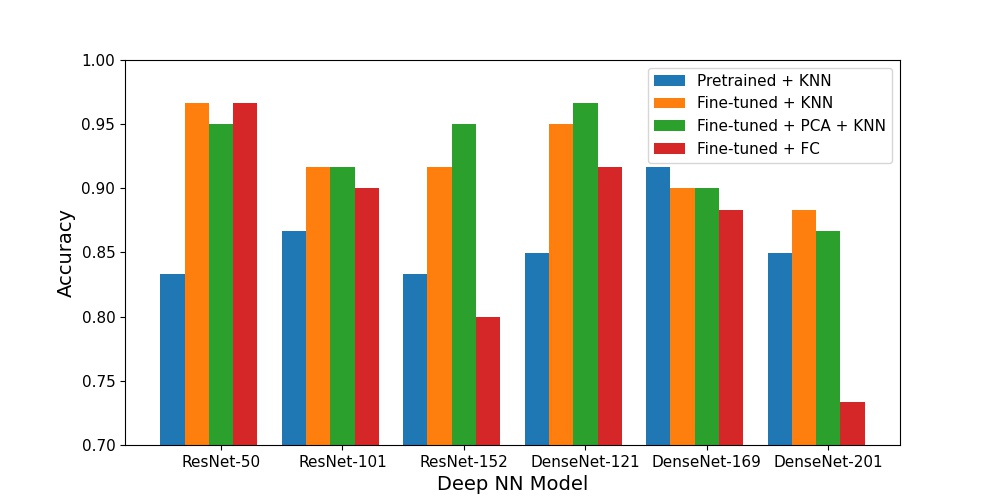}
      \caption{Classification performance using different combinations of features extractors (pretrained deep NN models, fine-tuned deep NN models, and fine-tuned deep NN models with PCA dimension reduction) and classifiers (k-NN-based classifiers and fine-tuned FC layer classifier).}
      \label{comparison}
\end{figure}

First, we use the pre-trained deep NN models as the feature extractor, and use a k-NN classifier to predict the abdominal organ in the image. After that, we use the fine-tuned NN models without the final layer as the feature extractor, and evaluate the classification accuracy of both the k-NN classifier and the fine-tuned FC layer classifier. Finally, we conducted PCA dimension reduction on the features extracted by the fine-tuned NN models for k-NN classification. In each experiment, the Euclidean distance, City block distance, Canberra distance and Cosine distance are used as distance metrics to measure the similarity between image features, and the value of k is optimized for each configuration. The final results are shown in Table \ref{result}, where the best accuracy for each deep NN architecture when using the same distance metrics in k-NN classification is indicated in bold, and the best accuracy among all the configurations is indicated in red. 

It can be seen that the best classification accuracy ($96.67\%$) is achieved by using ``fine-tuned ResNet-50 feature extractor + k-NN classifier", ``fine-tuned ResNet-50 + FC layer classifier", and ``fine-tuned DenseNet-121+PCA feature extractor + k-NN classifier". Despite that the FC layer classifier reaches a high classification accuracy when used with the ResNet-50 model, for all the other NN models, the fine-tuned FC layer classifier is outperformed by the k-NN classification based method. Also, it is observed that in most cases, the feature extractors with PCA achieve the best performance over the feature extractors without using PCA, showing that the dimension reduction method can effectively preserve the important features for better classification results. It is found that the choice of distance metrics does not have a great influence on the classification performance on the test data.

In order for a clear comparison of different feature extractors and classifiers, we compare the highest accuracy achieved among four distance metrics when using different combination of feature extractors and classifiers, as shown in Fig. \ref{comparison}. When using pretrained NN feature extractors combined with the k-NN classifier, the best accuracy is achieved by models with the medium depth (ResNet-101 and DenseNet-169). When using fine-tuned NN models as the feature extractor combined with the k-NN classifier, the accuracy improves by a large margin compared with the pretrained models, and it is interesting to see that a shallower network architecture yields a better performance. This may be due to the overfitting of deeper networks to the relatively small training dataset used in this work. It is also observed that doing PCA on the features extracted by the fine-tuned NN models can further improve the classification accuracy over the methods without doing PCA in most cases. This is mainly because PCA can get rid of correlated features and reduce over-fitting to the training data. In addition, as shown in both Table \ref{result} and Fig. \ref{comparison}, the k-NN based classification can yield better results compared with the fine-tuned FC layer classifier for all the NN models except ResNet-50. 

For the evaluation of the real-time performance, we further report the inference time of the k-NN-based classification and FC layer based classification. The average processing time for one image achieved on an AMD Ryzen 9 5950X CPU for the abdominal organ recognition task using the pretrained or fine-tuned deep NN feature extractor and k-NN classifier is $0.10s$, and that for the methods using fine-tuned deep NN feature extractor and FC layer classifier is $0.07s$. The results show that our method can generally satisfy the requirements for real-time recognition of abdominal organs in routine ultrasound scans.

\begin{figure}[t]
      \centering
      \includegraphics[scale=1.0,angle=0,width=0.49\textwidth]{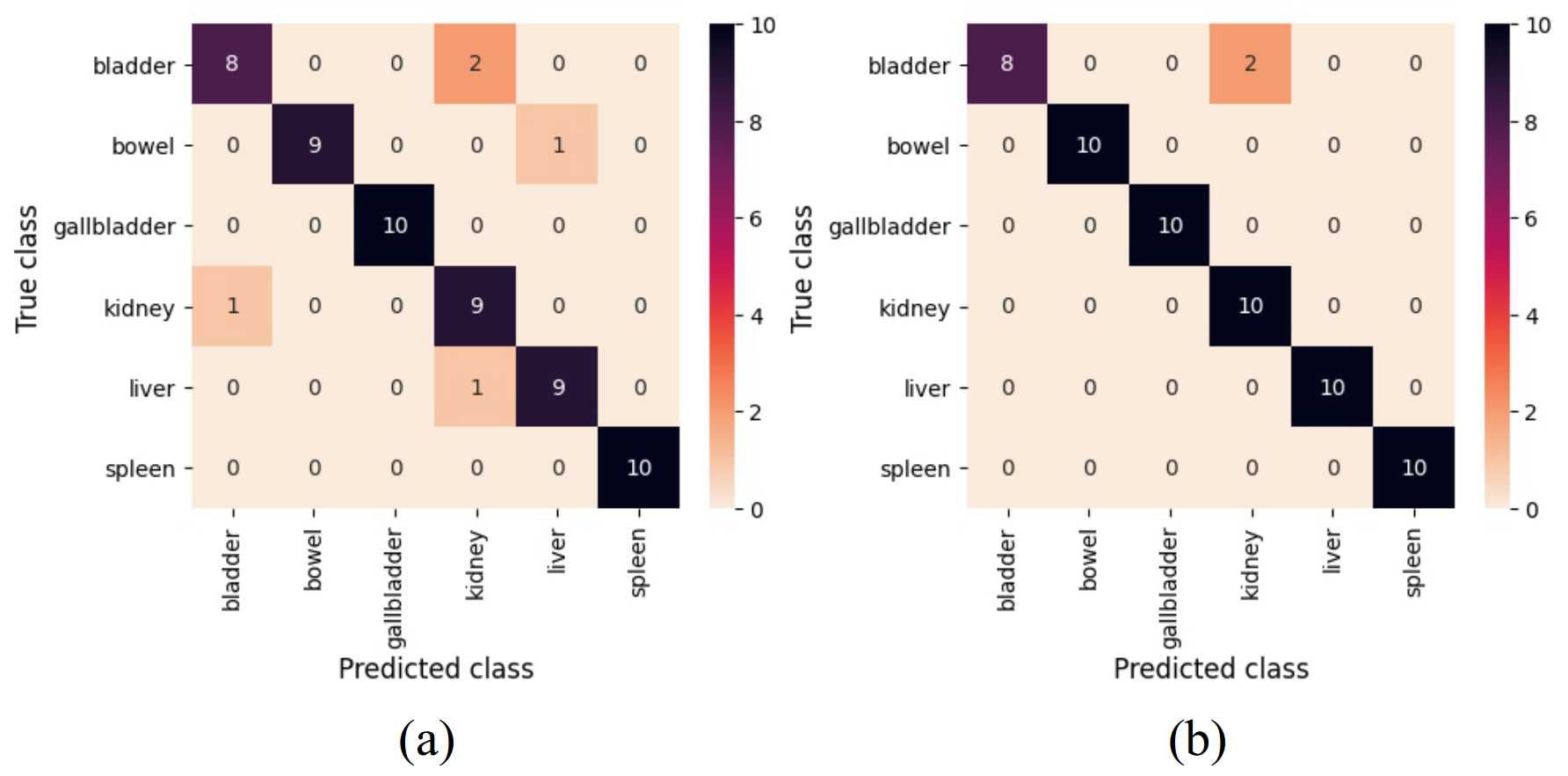}
      \caption{Confusion matrices of the methods (a) ``fine-tuned DenseNet-121 feature extractor + FC layer classifier" and (b) ``fine-tuned DenseNet-121+PCA feature extractor + k-NN classifier" for the classification of abdominal ultrasound images in the test set.}
      \label{confusion}
\end{figure}

\subsection{Qualitative Evaluation}

\begin{figure}[t]
      \centering
      \includegraphics[scale=1.4,angle=0,width=0.47\textwidth]{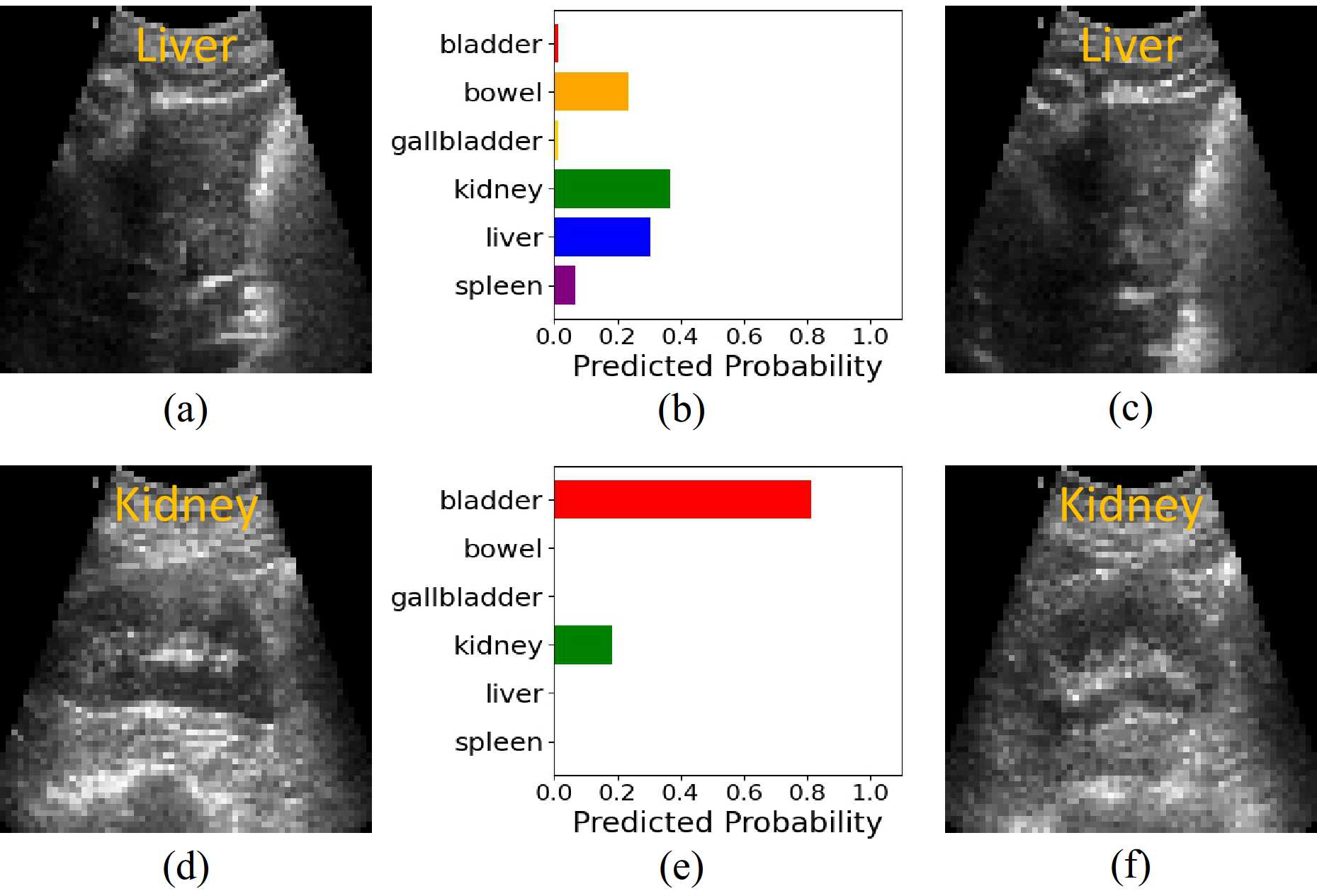}
      \caption{Qualitative evaluation of our method for abdominal ultrasound image classification. (a) and (d) show two test images with ground truth labels (yellow), which are misclassified by the ``fine-tuned DenseNet-121+PCA feature extractor + FC layer classifier". (b) and (e) show the corresponding predicted probability by the FC layer for each test image. (c) and (f) show the closest image in the database found by our method ``fine-tuned DenseNet-121+PCA feature extractor + k-NN classifier", with the labels indicated in yellow.}
      \label{qualitative}
\end{figure}

We further investigate the effectiveness of our method through qualitative analysis. We take a close look at the performance of the ``fine-tuned DenseNet-121 feature extractor + FC layer classifier" and ``fine-tuned DenseNet-121+PCA feature extractor + k-NN classifier" (k=3, using city block distance) on the test images. The confusion matrices of the two methods are illustrated in Fig. \ref{confusion}(a-b), respectively. It can be observed that a total of $5$ images are misclassified by the first method, of which $2$ images are misclassified by the second method. As shown in Fig. \ref{qualitative} (a) and (d), two test images that are misclassified by the FC layer classifier-based method but correctly recognized by the k-NN-based method are illustrated for a detailed discussion. The ground-truth labels of the two images are ``liver" and ``kidney", respectively. However, the FC layer classifier mistakenly recognizes the liver image as ``kidney", with a predicted probability (green) slightly higher than that of the true class (blue), as shown in Fig. \ref{qualitative}(b). In contrast, our method does not directly predicts the organ in the image by the network, but uses the fine-tuned deep NN as feature extractor and uses a k-NN classifier to find the closest image in the database. We can observe that our k-NN-based method successfully retrieves a liver image from the database to correctly predict the organ in the test image, as shown in Fig. \ref{qualitative}(c). 

For the recognition of the kidney image in Fig. \ref{qualitative}(d), it can be seen that the FC layer classifier-based method predicts the abdominal organ in the image as ``bladder" with a high confidence. This may be because of the large shadowed area in the image center, which looks a bit similar with the bladder images (see Fig. \ref{organs}(a)) and results in the confusion of the network. However, our method picks a kidney image from the database which has the closest distance in the feature space with the test data, and correctly predicts the organ in the test image as ``kidney", as shown in Fig. \ref{qualitative}(f). The qualitative results demonstrate that the k-NN classification-based method can effectively deal with some cases that are challenging to classify by directly applying a fine-tuned deep NN model.

\section{CONCLUSIONS}
In this paper, we propose a method that combines the deep neural networks with k-NN classification to automatically recognize six abdominal organs in the ultrasound images. Two state-of-the-art deep NN architectures, namely, the ResNet and DenseNet models with different depths, are pretrained and fine-tuned on our dataset to extract the image features, and PCA is performed for dimension reduction to remove the redundant information. The k-NN classifier is used with different distance metrics to predict the abdominal organs in the images. The experiment results demonstrate that with minimal training effort, our method is able to ``lazily" recognize the abdominal organs in the ultrasound images in real time, and outperforms the FC layer-based classification. In the future, this work can be applied in routine abdominal ultrasound examinations to assist in the diagnosis of various abdominal organs and reduce the workload of clinicians. It also has the potential to facilitate autonomous robotic ultrasound acquisitions, and can be easily generalized to the ultrasound imaging of other human tissues.

\addtolength{\textheight}{-1cm}   




%
%

\bibliographystyle{IEEEtran}   
\bibliography{root}

\end{document}